\documentclass[runningheads]{llncs}

\usepackage{eccv}

\usepackage{eccvabbrv}

\usepackage{graphicx}
\usepackage{amsmath}
\usepackage{amssymb}
\usepackage{booktabs}
\usepackage{array}
\usepackage{multirow}
\usepackage{color}
\usepackage{colortbl}
\usepackage{framed}
\usepackage{bm}
\usepackage{bbm}
\usepackage{xspace}
\usepackage{enumitem}
\usepackage{cite}
\usepackage{xparse}
\usepackage{xcolor}
\usepackage{algorithm}
\usepackage{url}
\usepackage{algorithmic}
\usepackage{pifont}
\usepackage{bbding}
\usepackage{mathtools}

\usepackage{pgfplots}
\usepackage{pgfplotstable}
\pgfplotsset{compat=1.18}

\usepackage[normalem]{ulem}
\useunder{\uline}{\ul}{}

\definecolor{Gray}{gray}{0.9}
\definecolor{LightCyan}{rgb}{0.88,0.95,1}
\definecolor{blond}{rgb}{0.98, 0.94, 0.75}

\def \ie {\emph{i.e.}}

\def \etal {\emph{et al.}}

\newcommand{\tit}[1]{\smallbreak\noindent\textbf{#1.}}
\newcommand{\tinytit}[1]{\noindent\textbf{#1.}}

\usepackage[accsupp]{axessibility}  

\usepackage{hyperref}

\usepackage{orcidlink}

\begin{document}

\title{Personalizing Multimodal Large Language Models for Image Captioning: An Experimental Analysis} 

\titlerunning{Personalizing Multimodal Large Language Models for Image Captioning}

\author{Davide Bucciarelli\inst{1}\orcidlink{0009-0002-9652-8311} \and
Nicholas Moratelli\inst{1}\orcidlink{0000-0001-9362-5680} \and\\
Marcella Cornia\inst{1}\orcidlink{0000-0001-9640-9385} \and
Lorenzo Baraldi\inst{1}\orcidlink{0000-0001-5125-4957} \and
Rita Cucchiara\inst{1,2}\orcidlink{0000-0002-2239-283X}}

\authorrunning{D. Bucciarelli et al.}

\institute{University of Modena and Reggio Emilia, Italy \\
\and
IIT-CNR, Italy\\
\email{\{name.surname\}@unimore.it}}
\maketitle

\sloppy

\begin{abstract}
The task of image captioning demands an algorithm to generate natural language descriptions of visual inputs. Recent advancements have seen a convergence between image captioning research and the development of Large Language Models (LLMs) and  Multimodal LLMs -- like GPT-4V and Gemini -- which extend the capabilities of text-only LLMs to multiple modalities. This paper investigates whether Multimodal LLMs can supplant traditional image captioning networks by evaluating their performance on various image description benchmarks. We explore both the zero-shot capabilities of these models and their adaptability to different semantic domains through fine-tuning methods, including prompt learning, prefix tuning, and low-rank adaptation. Our results demonstrate that while Multimodal LLMs achieve impressive zero-shot performance, fine-tuning for specific domains while maintaining their generalization capabilities intact remains challenging. We discuss the implications of these findings for future research in image captioning and the development of more adaptable Multimodal LLMs.
\end{abstract}

\section{Introduction}
\label{sec:intro}
The task of image captioning requires an algorithm to describe a visual input in natural language. Over the last years, researchers have made remarkable progress in developing approaches specifically devoted to image description, with the aim of increasing visual encoding capabilities~\cite{vinyals2015show,anderson2018bottom}, finding proper architectures and multimodal connectors~\cite{barraco2023little,li2022comprehending}, and improving linguistic fluency, relevance, and adherence to a desired description style~\cite{moratelli2024revisiting,kornblith2023guiding,dessi2023cross}. These advancements have not only enhanced the ability of models to generate accurate and contextually appropriate captions but have also contributed to bridging the gap between visual understanding and language generation.

Given the inherent multimodal nature of the task, the evolution of image captioning research has many times intersected that of Large Language Models (LLMs)~\cite{ramos2023smallcap,mokady2021clipcap,kornblith2023guiding} and, more recently, it is crossing that of Multimodal LLMs (MLLMs)~\cite{li2023blip,alayrac2022flamingo,wang2023cogvlm,caffagni2024wiki}, which are their natural multimodal extension. The surge of sophisticated text-only LLMs~\cite{brown2020language,touvron2023llama,vicuna2023}, and particularly their capacity for in-context learning, has indeed encouraged researchers to broaden the scope of these models to encompass multiple modalities, both as inputs and outputs. This expansion has led to the development of cutting-edge models such as GPT-4V~\cite{achiam2023gpt} and Gemini~\cite{team2023gemini}, which showcase state-of-the-art performance in various multimodal tasks and applications.

A significant example of the interplay between image captioning research and large multimodal models can be found by analyzing the evolution of their training methodologies. Starting from 2015, captioners have been fine-tuned with reinforcement learning objectives to maximize non-differentiable metrics like CIDEr~\cite{vedantam2015cider}, \ie~through self-critical sequence training (SCST)~\cite{rennie2017self}. This approach, although conducted on a smaller scale, closely resembles that of the reinforcement learning from human feedback (RLHF) paradigm~\cite{ouyang2022training}, which has been a fundamental tool to develop instruction-aligned LLMs and, ultimately, increase their utility in real-world scenarios. In RLHF, indeed, the LLM is fine-tuned to align itself to a trained reward model -- replace the trained reward with a non-differentiable metric, and you immediately get a fine-tuning strategy that is conceptually equivalent to SCST. Coming to MLLMs, the similarities between the two tasks are evident, with research on MLLMs questioning the best way to fuse visual features into a Transformer decoder~\cite{tong2024cambrian,tong2024eyes,caffagni2024r} -- a question that image captioning literature has been tackling several times in the past.

Considering this overlap in technical goals, the recent surge of MLLMs and the variety of multimodal tasks that they can perform, a natural question arises: are MLLMs the definite replacement for image captioning networks? In this paper, we contribute to finding an answer to this question, by analyzing the performance of different MLLMs on multiple image description benchmarks. In addition to investigating the zero-shot performance of pre-trained models in comparison with that of a state-of-the-art captioner, we also move a step forward and test the adaptation capabilities of MLLMs when it comes to adhering to the classical description style of captioners, which is very concise, grammatically correct, and focuses on everyday objects. Also, we test whether this adaptation can still maintain the generalization capabilities of the MLLM and work well on other semantic domains.

To test this adaptation -- or, better to say, \textit{personalization} -- capabilities of MLLMs, we employ different fine-tuning strategies, ranging from full fine-tuning to a wide range of parameter-efficient fine-tuning (PEFT) techniques, including prompt learning~\cite{lester2021power}, prefix tuning~\cite{li2021prefix}, low-rank adaptation~\cite{hu2021lora}, and weight-decomposition in low-rank adaptation~\cite{liu2024dora}. By assessing the performance of current MLLMs for image description, and their adaptation capabilities to different semantic and description domains, we aim to provide a comprehensive evaluation of whether these models can truly replace specialized image captioning networks. Our findings reveal that, while MLLMs exhibit strong zero-shot performance across various benchmarks, their adaptability to specific description styles through fine-tuning is still an open challenge. We conclude by discussing the implications of our results for future research directions in both image captioning and the development of more versatile and adaptive MLLMs.

\section{Related Work}
\label{sec:related}
\tinytit{Standard Image Captioning}
Early efforts in image captioning primarily focused on detecting key objects within a scene to populate predefined templates~\cite{socher2010connecting,yao2010i2t}. Subsequent research evolved to employ RNN-based encoder-decoder architectures, where visual input was encoded using a CNN and then exploited to condition the generation process through an RNN~\cite{vinyals2015show,karpathy2015deep}. These methodologies were further refined with the introduction of attention-based strategies, which applied attention mechanisms to either spatial regions~\cite{anderson2018bottom} or semantic graphs~\cite{yang2019auto}. Recently, Transformer-based architectures have emerged as the standard in image captioning~\cite{huang2019attention,cornia2020smart,cornia2022explaining}, often in combination with CLIP-based~\cite{radford2021learning} visual features which demonstrate increased semantics leading to better performance~\cite{barraco2022unreasonable,li2022comprehending,barraco2023little}. Despite being a well-established task in literature, it has historically struggled with generalization and tends to produce very literal captions. In this regard, recent approaches have proposed fine-tuning strategies guided by open-vocabulary metrics~\cite{sarto2023positive,hessel2021clipscore,sarto2024bridge} to enhance the descriptive capacity of the models~\cite{kornblith2023guiding,cho2022fine,moratelli2024revisiting}.

\tit{Image Captioning with Multimodal LLMs}
In the last year, MLLMs have become predominant in performing a wide range of vision-and-language tasks including visual dialogue, image description, and visual question answering~\cite{caffagni2024r}. Almost all existing MLLMs, indeed, adopt large-scale architectures to tackle the challenge of bridging visual and language modalities, connecting a pre-trained LLM with a large-scale visual encoder (\ie~typically CLIP or its variants). 

MLLMs can be categorized considering the type of multimodal connections they employ. Following the widely-used LLaVA model family~\cite{liu2023visual,liu2023improved,liu2024llavanext}, the prevalent strategy in this domain involves using an MLP~\cite{zhao2023svit,wang2023cogvlm} or a single linear layer~\cite{chen2023minigpt,lin2023sphinx} to establish multimodal connections. Several variations have been introduced, such as LLaMA-Adapter~\cite{gao2023llama} that proposes an alternative attention mechanism with zero gating, and the approach introduced by Cha~\etal~\cite{cha2023honeybee} that replaces linear layers with convolutions. Another significant category of models is built upon Q-Former architecture introduced in~\cite{li2023blip}. On this line, mPLUG-Owl~\cite{ye2023mplug} streamlines Q-Former by incorporating a visual abstractor component that condenses visual information into distinct trainable tokens. Similarly, Qwen-VL~\cite{bai2023qwen} employs a single-layer cross-attention module with learnable queries to compress visual features. Other approaches integrate dense cross-attention blocks within the existing pre-trained layers of the LLM~\cite{awadalla2023openflamingo,alayrac2022flamingo}. This method is often used in conjunction with a Perceiver model~\cite{jaegle2021perceiver}, reducing the number of visual tokens before their integration into the language model.

Despite their rapid evolution, the performance analysis of MLLMs in image captioning remains significantly under-explored. Only a few MLLMs are directly trained and evaluated on this task using standard benchmarks, while others treat image description as an inherent capability. On a different line, some recent studies~\cite{liu2023aligning, wang2023vigc,yin2023woodpecker} have started to estimate the hallucination degree of MLLMs, a crucial aspect in this domain given the level of detail they can generate even when describing input images. Unlike existing literature, this paper aims to analyze standard MLLMs when generating image descriptions and explore how they can be better adapted to the task by comparing different fine-tuning techniques.

\tit{Parameter Efficient Fine-tuning Techniques}
Adapting LLMs to a specific task may prove impractical due to the substantial computational resources required for complete fine-tuning. In such scenarios, the adoption of PEFT techniques represents a feasible alternative. The principal strategies include (i) prompt-tuning that entails learning a small set of vectors, used as soft prompts fed into the model before the input text~\cite{hambardzumyan2021warp,lester2021power,li2021prefix,liu2023gpt,moratelli2024learnable}; (ii) LoRA~\cite{hu2021lora}, where pre-trained model weights remain frozen while introducing trainable rank decomposition matrices into each layer; (iii) QLoRA~\cite{dettmers2023qlora}, designed to reduce the memory footprint of LLMs while preserving full 16-bit fine-tuning task performance and (iv) DoRA~\cite{liu2024dora} that decomposes a pre-trained model into magnitude and directional components, utilizing LoRA for directional adjustments, thereby efficiently reducing the count of trainable parameters. Despite the availability of a diverse range of techniques, to the best of our knowledge, there has been no experimental analysis conducted to compare them. This paper investigates the impact of PEFT optimization on model performance when customizing the MLLM for a specific task (\ie~that of image captioning).

\section{Proposed Method}
\label{sec:method}
\subsection{Preliminaries}
An MLLM usually takes as input a multimodal input, comprising both image and text, and generates a textual output in an autoregressive manner. Formally, the architecture is trained to model a probability distribution $p(w_t|I, w_0, w_1, ..., w_{t-1}, \theta)$, where $\theta$ denotes the parameters of the model, $I$ represents an input image, and ${w_0,..,w_{t-1}}$ denotes the textual prompt. The textual prompt usually includes a pre-defined system-level prompt and a question related to the input image, given by the user. Clearly, a standard MLLM can only rely on the user prompt, the input image, and the knowledge stored in its internal parameters (\ie~$\theta$) to accommodate requests. 

In the rest of the paper, we employ LLaVA~\cite{liu2023visual} as our reference MLLM. LLaVA exploits the capabilities of a pre-trained LLM (\ie~Vicuna~\cite{vicuna2023}) and a pre-trained visual model (\ie~a CLIP-based visual encoder~\cite{radford2021learning}), which are interconnected through an MLP adapter, in charge of converting CLIP features to dense input tokens. For an input image $I$, therefore, LLaVA utilizes a pre-trained CLIP visual encoder $E_v$, extracts a dense grid of visual features $Z_v = E_v(I)$, which is then projected via a learnable MLP to produce a sequence of dense embedding tokens $v_o, v_1, ..., v_N$. Finally, these are prepended to the system prompt, and the full sequence of visual and textual tokens is then given as input to the LLM component of the model.

\begin{figure}
    \centering
    \includegraphics[width=0.98\linewidth]{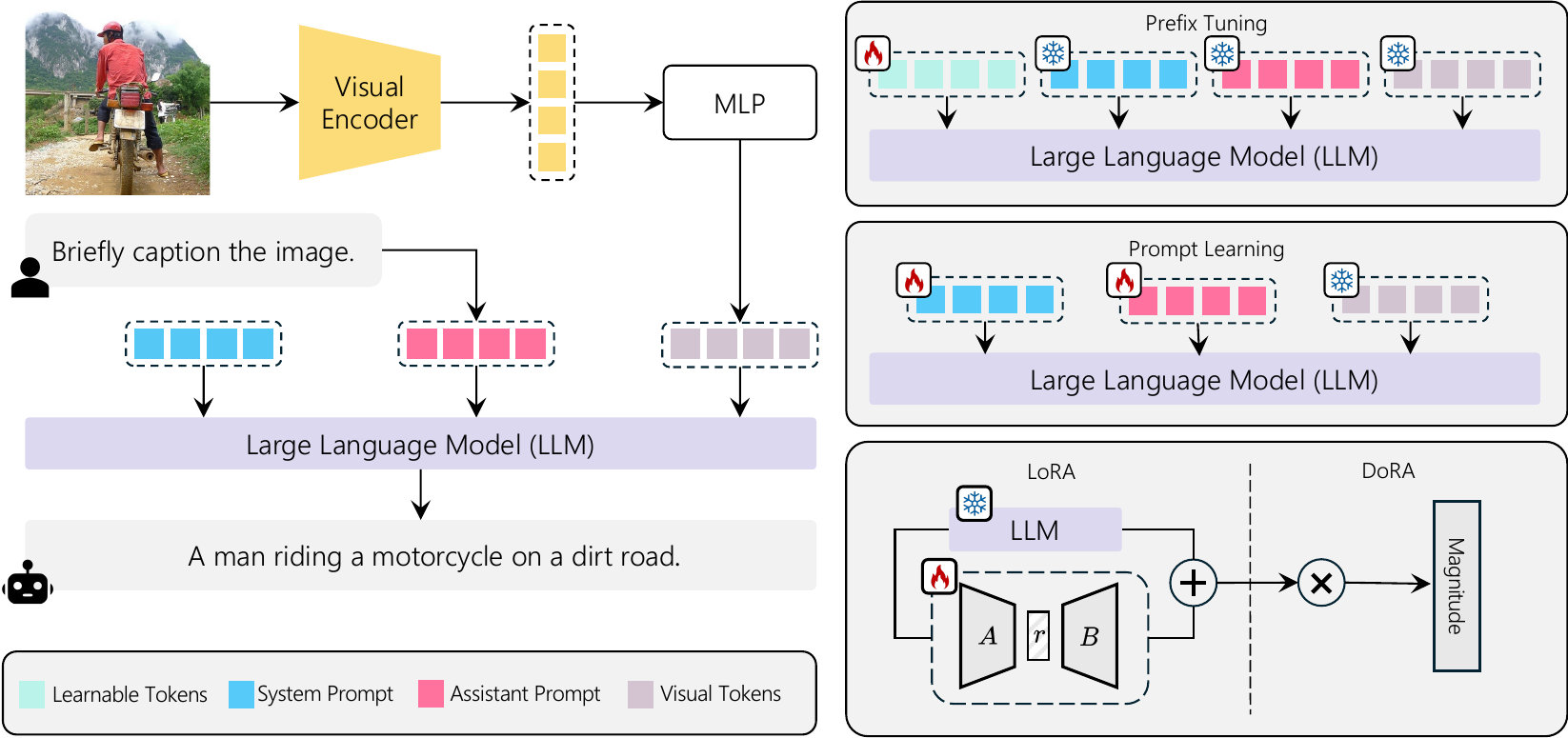}
    \vspace{-0.15cm}
    \caption{Overview of our approach. We investigate whether Multimodal LLMs can supplant traditional captioners by assessing their adaptability to different semantic domains and through the usage of different adaptation techniques.}
    \label{fig:strategies}
    \vspace{-0.2cm}
\end{figure}

\subsection{Personalization Strategies}\label{sec:personalization}
To adapt MLLMs to specific description styles and semantic domains, we investigate several parameter-efficient fine-tuning (PEFT) techniques. 

\tit{Prompt Learning}
To adapt the MLLM to perform classical image captioning, the most straightforward option is to enrich the input context by injecting learnable vectors into its embedding. This is usually done by adding \textit{new} embedding vectors to an existing prompt, which are initialized from scratch and trained through stochastic gradient descent. In our preliminary experiments, however, we found it beneficial to fine-tune the user prompt and the system prompt embeddings, rather than injecting new embeddings which might be more complex to initialize. Formally, the distribution of the MLLM is conditioned on visual tokens, a system prompt, and a trainable user prompt, leading to
\begin{equation}
    p(w_t|
    \overbracket{v_o, v_1, ..., v_N}^{\text{Visual tokens}},\ \ \ \ \underbracket{w_0, w_1, ..., w_{t-1}}_\text{\textcolor{red}{Learnable System prompt}}, \overbracket{e_0, e_1, ..., e_{\tau}}^\text{\textcolor{blue}{Learnable User prompt}}),    
\end{equation}
where $e_0, ..., e_{\tau}$ represents the trainable embeddings of the user prompt. The set of trainable parameters $\theta^\ast$, in this case, is simply $\theta^\ast = \{e_0, ..., e_{\tau}\}$. Differently from the standard formulation of MLLMs, by fine-tuning a portion of the input context, we allow the model to generate more specific answers.

\tit{Prefix Tuning}
Differently from the previous case, in this case we add a sequence of learnable embeddings to every layer of the Transformer decoder of the MLLM. While this formulation does not allow a straightforward meaningful initialization of the embeddings, like in the case of prompt learning, it comes with the advantage of injecting trainable knowledge at different layers of the architecture, which might increase the degree of adaptation of the model.
Formally, the input embeddings of the $i$-th layer are adapted as
\begin{equation}
    \bm{h}^i = \left[
    \underbracket{h_0^i, h_1^i, ..., h_T^i}_{\text{Regular input embeddings}},\ \ \ \ \underbracket{e_0^i, e_1^i, ..., e_{\tau}^i}_\text{\textcolor{blue}{Learnable embeddings}} \right],
\end{equation}
where $e_0^i, ..., e_{\tau}^i$ represents the trainable embeddings of a given layer. In this case, the set of trainable parameters $\theta^\ast$ is defined as $\theta^\ast = \bigcup_{i=1}^L \{e_0^i, ..., e_{\tau}^i\}$, where $L$ represents the number of layers of the MLLM.

\tit{LoRA}
We now turn to a different approach to adaptation, where instead of adding learnable tokens or embeddings, either at the input layer or at every layer, we aim at fine-tuning all the weights $\theta$ of the architecture. To constrain the computational complexity of the adaptation, and keep a safe regularization against overfitting, we fine-tune a low-rank adaptation of weight matrices~\cite{hu2021lora} instead of directly performing a full fine-tuning. 

Without loss of generality, in the following, we describe our approach for the case of a fully-connected layer, which are a key ingredient of many Transformer-based models as they build up the attention operator. Given a pre-trained layer $f$, with weight $W_0 \in \theta$, $W_0 \in \mathbb{R}^{d \times k}$ and bias $b \in \theta$, which applies a transformation $f(x) = x W_0^{\intercal} + b$ to its input tensor $x \in \mathbb{R}^k$, we re-parametrize its transformation during the training phase by adding a low-rank trainable component $\tilde{W}$, initialized from zero. We then fine-tune only the low-rank decomposition, leaving the rest of the layer frozen. Formally,
\begin{equation}
    \label{eq:decomp}
    f(x) = x\underbracket{W_0^\intercal}_\text{\textcolor{blue}{\SnowflakeChevron}} + x\overbracket{\tilde{W}^\intercal}^{\ast} + \underbracket{b}_\text{\textcolor{blue}{\SnowflakeChevron}}. \text{ with } \tilde{W} = BA, 
\end{equation}
where $A$ and $B$ provide a bottleneck that creates a low-rank decomposition which is trainable (denoted with $\ast$, above), with $B \in \mathbb{R}^{d \times r}$, $A \in \mathbb{R}^{r \times k}$ and $r$ is the rank of the decomposition. During fine-tuning, $W_0$ and $b$ are kept frozen (\textcolor{blue}{\small{\SnowflakeChevron}}) and we backpropagate gradient only on $A$ and $B$. These are respectively initialized with a Gaussian and zero initialization, so that, at the beginning of fine-tuning, $\tilde{W} = BA$ is a zero matrix and $f$ behaves exactly as in the pre-trained state. 

We apply the low-rank re-parametrization to all fully connected layers of the MLLMs. The set of trainable parameters $\theta^\ast$ is, therefore, defined as $\theta^\ast = \bigcup_{i} \{A_i, B_i \}$, where $i$ runs on all fully-connected layers of the network.

\tit{DoRA}
As an alternative to the low-rank adaptation mentioned above, we also investigate a weight-decomposed low-rank adaptation (DoRA)~\cite{liu2024dora}, which is known in the literature for outperforming LoRA also when fine-tuning MLLMs such as LLaVa~\cite{liu2023visual}. Specifically, DoRA initially decomposes the pre-trained weight $W_0$ into its magnitude and directional components, then fine-tunes both of them. Because the directional component is larger in terms of parameter numbers, it fine-tunes it using a LoRA decomposition.

Formally, given a pre-trained weight $W_0 \in \mathbb{R}^{d \times k}$ from a layer $f$, the matrix weight is fine-tuned as
\begin{equation}
    \label{eq:decomp2}
    f(x) = x \cdot \overbracket{m}^{\ast} \frac{\overbracket{W_0}^{\text{\textcolor{blue}{\SnowflakeChevron}}} + \overbracket{BA}^{\ast}}{\| \underbracket{W_0}_{\text{\textcolor{blue}{\SnowflakeChevron}}} + \underbracket{BA}_{\ast} \|},
\end{equation}
where $m$ indicates a trainable magnitude component and, again, the $\ast$ notation indicates the only trainable weights. Clearly, DoRA adds an additional term to the set of trainable weights for the overall architecture, leading to $\theta^\ast = \bigcup_{i} \{m_i, A_i, B_i \}$, where $i$ runs on all fully-connected layers of the network.

\section{Experimental Evaluation}
\label{sec:experiments}
\subsection{Datasets and Evaluation Metrics}
In our experiments, we employ a set of five commonly used datasets for the image captioning task, namely COCO~\cite{lin2014microsoft}, nocaps~\cite{agrawal2019nocaps}, CC3M~\cite{sharma2018conceptual}, VizWiz~\cite{gurari2020captioning}, and TextCaps~\cite{sidorov2020textcaps}. All our training experiments are conducted on COCO, while evaluations are reported on all considered datasets. For nocaps, CC3M, VizWiz, and TextCaps, we report the results on the validation set of each dataset.

\tit{COCO~\cite{lin2014microsoft}} It is the standard dataset for the task and contains 
more than 120,000 images, each of them annotated with five different captions. In our experiments, we follow the splits provided by Karpathy~\etal~\cite{karpathy2015deep}, where 5,000 images are reserved for the validation set and 5,000 for the test set.

\tit{nocaps~\cite{agrawal2019nocaps}} This dataset consists of 15,100 elements from the Open Images~\cite{kuznetsova2018open} validation and test sets, annotated with human-annotated captions. Images are divided into validation and test splits, respectively with 4,500 and 10,600 images.

\tit{CC3M~\cite{sharma2018conceptual}} It is a large-scale image captioning dataset composed of roughly 3.3 million images sourced from the web. The validation set is composed of approximately 14,000 elements. Each image is paired with an alt-text description, that usually focuses on the main concept of the image.

\tit{VizWiz~\cite{gurari2020captioning}} This dataset aims to test the ability of image captioning models to assist blind people. It features 39,000 images taken by visually impaired users, each paired with five ground-truth captions. Images are grouped into training, validation, and test sets with 23,431, 7,750, and 8,000 elements each.

\tit{TextCaps~\cite{sidorov2020textcaps}} It includes over 145,000 captions for more than 28,000 images, where each caption requires understanding and interpreting the textual content present in the image. The images are split into training, test, and validation sets, with respectively 21,953, 3,166, and 3,289 elements.

\tit{Evaluation} For what concerns the evaluation metrics, we employ standard captioning scores, namely BLEU-4~\cite{papineni2002bleu}, METEOR~\cite{banerjee2005meteor}, ROUGE~\cite{lin2004rouge}, CIDEr~\cite{vedantam2015cider}, and SPICE~\cite{anderson2016spice}. Additionally, we report the results in terms of CLIP-Score~\cite{hessel2021clipscore} that does not rely on ground-truth captions and tends to favor a high degree of descriptiveness at the expense of grammatical fluency and correctness~\cite{moratelli2024revisiting,cho2022fine}.

\subsection{Implementation and Training Details}
In our experiments, we focus on the smallest versions of the LLaVA models, selecting LLaVA-v1.5-7B~\cite{liu2023improved} and LLaVA-v1.6-7B~\cite{liu2024llavanext}, both equipped with the Vicuna-7B~\cite{vicuna2023} language model and CLIP ViT-L/14@336~\cite{radford2021learning} as visual encoder. To fine-tune the considered MLLMs, we utilize the standard token-level cross-entropy loss for all training experiments. Unless otherwise specified, the CLIP-based visual encoder, the MLP adapter, and the LLM layers are kept frozen.

Both considered MLLMs are trained using the personalization strategies described in Sec.~\ref{sec:personalization}. All results are compared against the original MLLM (\ie~the MLLM without any fine-tuning stages), tested in zero-shot using a fixed prompt\footnote{In our experiments, we employ ``\texttt{Briefly caption the image}'' as input prompt.} to generate the output image description. For prompt learning, we implement a slight variation of the usual framework. Specifically, we initialize the learnable prompt with the same sentence used for the original MLLM, concatenated with the standard system prompt of the model, which is then optimized during training. This results in 48 learnable tokens. In prefix tuning, we employ a fixed prompt identical to the one used for the zero-shot evaluation and concatenate a single learnable token at the beginning of the input for each decoder layer of the LLM, trimming the extra output token at each layer. For LoRA and DoRA, we use identical parameters (\ie~the rank $r$ is set to $128$ and the scaling parameter $\alpha$ is equal to $256$) and keep the same input prompt used in the other settings.

To ensure fair comparison among different fine-tuning methods, we maintain a consistent training setup across all tests. In particular, all training experiments are performed on a single node with four 64GB NVIDIA A100 GPUs. During each training phase, the model undergoes four epochs on the COCO dataset, with the model exhibiting the lowest validation loss being selected at the end of each run. We use a batch size of 32 for all experiments, employing gradient accumulation steps as needed for memory constraints. The standard SGD optimizer is utilized, along with a cosine learning rate scheduler, with a maximum learning rate equal to $2\times10^{-2}$ and a minimal value of $1\times10^{-5}$.

\subsection{Experimental Results} 
As previously mentioned, we consider two existing MLLMs (\ie~LLaVA-v1.5 and LLaVA-v1.6) and evaluate their performance across different captioning datasets. In addition to the results using the original model, we fine-tune each MLLM using four different personalization strategies, namely prompt learning, prefix tuning, LoRA, and DoRA (cf. Sec.~\ref{sec:personalization}). Moreover, we report the results of each MLLM fine-tuned by directly optimizing all parameters of both the vision-to-language adapter and all layers of the LLM. To have a direct comparison with a standard captioning model not based on MLLMs, we also consider the performance of a standard Transformer-based encoder-decoder model trained from scratch on the COCO dataset. Specifically, we follow the architecture of the CLIP-Captioner proposed in~\cite{barraco2022unreasonable} and train it using the same visual encoder used in the considered LLaVA models (\ie~CLIP ViT-L/14@336). In the following, we first report the results on the standard COCO dataset, thus following an in-domain evaluation. Then, we analyze the generalization capabilities to out-of-domain datasets showing the results on nocaps, CC3M, VizWiz, and TextCaps.

\begin{table}[t]
\caption{In-domain results on the COCO dataset. Bold font indicates the best results for the same MLLM, while underline indicates the overall best scores.}
\vspace{-0.1cm}
\label{tab:coco-table}
\footnotesize
\centering
\setlength{\tabcolsep}{.35em}
\resizebox{0.82\linewidth}{!}{
\begin{tabular}{lccc cccccc}
\toprule
\textbf{Model} & & \textbf{PEFT} & & B-4 & M & R & C & S & CLIP-S \\
\midrule
CLIP-Captioner~\cite{barraco2022unreasonable} & & - & & . & \underline{29.9} & \underline{58.2} & \underline{126.2} & 22.6 & 0.752 \\  
\midrule
\multirow{6}{*}{LLaVA-v1.5-7B~\cite{liu2023improved}} && - & & 18.7 & 22.4 & 46.6 & 53.9 & 24.8 & \underline{\textbf{0.806}} \\
& & Prompt Learning & & 31.9 & 22.4 & 53.5 & 96.3 & 23.1 & 0.774 \\
& & Prefix Tuning & & 27.3 & 22.3 & 52.0 & 85.8 & 23.4 & 0.782 \\
& & LoRA & & 36.1 & 23.2 & 56.0 & 105.7 & 24.7 & 0.777 \\
& & DoRA & & 36.4 & 23.3 & 56.3 & 106.1 & 24.5 & 0.778 \\
& & Full Fine-tuning  & & \textbf{38.2} & \textbf{23.5} & \textbf{57.3} & \textbf{111.4} & \textbf{25.1} & 0.771 \\
\midrule
\multirow{6}{*}{LLaVA-v1.6-7B~\cite{liu2024llavanext}} && - & & 6.8 & 16.2 & 31.2 & 16.4 & 12.5 & 0.755 \\
& & Prompt Learning & & 33.0 & 22.8 & 54.5 & 100.0 & 24.1 & 0.774 \\
& & Prefix Tuning & & 25.5 & 19.4 & 48.4 & 74.1 & 19.7 & \textbf{0.784} \\
& & LoRA & & 36.9 & 23.3 & 56.6 & 108.6 & 24.8 & 0.771 \\
& & DoRA & & 36.1 & 23.1 & 56.1 & 106.4 & 24.7 & 0.777 \\
& & Full Fine-tuning & & \textbf{38.5} & \textbf{23.4} & \textbf{57.5} & \textbf{112.3} & \underline{\textbf{25.2}} & 0.774 \\
\bottomrule
\end{tabular}
}
\vspace{-0.3cm}
\end{table}

\tit{In-Domain Evaluation} Table~\ref{tab:coco-table} presents the in-domain results on the COCO dataset for the models under consideration. As indicated, the highest scores for each MLLM are highlighted in bold, while the overall best scores across all models and methods are underlined.

Firstly, examining the results of LLaVA-v1.5, it can be noticed that the highest scores are achieved by the full fine-tuning strategy with a CIDEr score of 111.4 points. Similar results are achieved by the LLaVA-v1.6 model where full fine-tuning again yields the highest scores in almost all metrics with a CIDEr score equal to 112.3 points. Among the other fine-tuning strategies, LoRA and DoRA are the ones that achieve the best results on both LLaVA versions. Overall, these results are not surprising: training a larger number of parameters using image-caption pairs from a specific dataset and evaluating the results on other pairs from the same dataset naturally leads to the best performance. This is further confirmed when taking into account the results achieved by the CLIP-Captioner, which are generally higher than all the others reported in the table. Since this model is trained from scratch on the COCO dataset, directly assessing the performance on the test set of the same dataset leads to the best scores on standard captioning metrics. 

Conversely, the best results in terms of CLIP-S are achieved by the original LLaVA-v1.5 model tested in a zero-shot manner on COCO. This underscores the descriptive capabilities of MLLMs which can generate highly detailed and usually long captions describing a given image. The descriptive style of common MLLMs, however, is far from the one present in standard captioning benchmarks like COCO which contains concise and timely descriptions, as demonstrated by the CIDEr scores of both zero-shot LLaVA-v1.5 and LLaVA-v1.6 models (\ie~53.9 and 16.4, respectively) which are significantly lower than those obtained by all fine-tuned versions.

\begin{table}[t]
\caption{Out-of-domain results on nocaps and CC3M datasets. Bold font indicates the best results for the same MLLM, while underline indicates the overall best scores.}
\vspace{-0.1cm}
\label{tab:out-domain-1}
\footnotesize
\centering
\setlength{\tabcolsep}{.22em}
\resizebox{\linewidth}{!}{
\begin{tabular}{lccc cccccc c cccccc}
\toprule
& & & & \multicolumn{6}{c}{\textbf{nocaps}} & & \multicolumn{6}{c}{\textbf{CC3M}} \\
\cmidrule{5-10} \cmidrule{12-17}
\textbf{Model} & & \textbf{PEFT} & & B-4 & M & R & C & S & CLIP-S & & B-4 & M & R & C & S & CLIP-S \\
\midrule
CLIP-Captioner~\cite{barraco2022unreasonable} & & - & & 7.3 & 15.9 & 32.4 
& 77.1 & 19.7 & 0.693 & & 1.9 & 9.0 & \underline{16.7} & \underline{29.1} & 9.5 & 0.651 \\  
\midrule
\multirow{6}{*}{LLaVA-v1.5-7B~\cite{liu2023improved}} && - & & 6.5 & \underline{\textbf{21.3}} & 31.2 & 61.8 & 23.6 & \underline{\textbf{0.793}} & & 1.2 & \underline{\textbf{11.0}} & 14.1 & 15.1 & 10.3 & \underline{\textbf{0.743}} \\
& & Prompt Learning & & 8.4 & 20.1 & 33.2 & 85.9 & 24.2 & 0.763 & & \underline{\textbf{2.0}} & 10.4 & \textbf{14.9} & \textbf{22.9} & 10.8 & 0.699 \\
& & Prefix Tuning & & \underline{\textbf{9.0}} & 20.7 & 33.3 & 84.1 & 24.6 & 0.772 & & 1.8 & 10.5 & 14.4 & 21.3 & \textbf{10.9} & 0.720 \\
& & LoRA & & 8.5 & 20.0 & 33.2 & 86.2 & 24.8 & 0.751 & & 1.7 & 10.1 & 14.3 & 21.6 & 10.6 & 0.714 \\
& & DoRA & & 8.5 & 20.2 & \textbf{33.4} & \textbf{87.3} & \textbf{24.9} & 0.758 & & 1.8 & 10.2 & 14.5 & 22.0	& 10.8 & 0.717 \\
& & Full Fine-tuning & & 8.7 & 20.1	 & 33.2 & 86.8 & 24.6 & 0.752 & &  1.7 & 10.0 & 14.3 & 20.8 & 10.4 & 0.720 \\
\midrule
\multirow{6}{*}{LLaVA-v1.6-7B~\cite{liu2024llavanext}} && - & & 2.3 & 14.8 & 19.6 & 18.1 & 10.8 & 0.749 & & 1.2	& 9.8 & 11.7 & 9.6 & 7.5 & 0.731 \\
& & Prompt Learning & & 8.8	& \textbf{20.4} & \underline{\textbf{34.1}} & \underline{\textbf{91.0}} & \underline{\textbf{25.5}} & 0.758 & & \textbf{1.8} & 10.1 & 14.4 & 20.7 & \underline{\textbf{11.0}} & 0.729 \\
& & Prefix Tuning & & 6.4 & 17.5 & 29.4 & 65.6 & 22.3 & 0.752 & & 1.4 & 9.7 & 12.1 & 15.1 & 8.6 & \textbf{0.738} \\
& & LoRA & & 8.8 & \textbf{20.4} & 33.7 & 89.8 & 25.3 & 0.762 & & \textbf{1.8} & \textbf{10.2} & 14.4 & 21.7 & 10.2 & 0.718 \\
& & DoRA & & \textbf{8.9} & \textbf{20.4} & 33.8 & \underline{\textbf{91.0}} & 25.4 & \textbf{0.763} & & \textbf{1.8} & \textbf{10.2} & \textbf{14.5} & \textbf{21.8} & 10.6 & 0.723 \\
& & Full Fine-tuning & & 8.5 & 20.1 & 33.2 & 86.6 & 24.6 & 0.753 & & 1.6 & 9.9 & 14.2 & 20.7 & 10.1 & 0.724 \\
\bottomrule
\end{tabular}
}
\vspace{-0.15cm}
\end{table}

\begin{table}[t]
\caption{Out-of-domain results on VizWiz and TextCaps datasets. Bold font indicates the best results for the same MLLM, while underline indicates the overall best scores.}
\vspace{-0.1cm}
\label{tab:out-domain-2}
\footnotesize
\centering
\setlength{\tabcolsep}{.22em}
\resizebox{\linewidth}{!}{
\begin{tabular}{lccc cccccc c cccccc}
\toprule
& & & & \multicolumn{6}{c}{\textbf{VizWiz}} & & \multicolumn{6}{c}{\textbf{TextCaps}} \\
\cmidrule{5-10} \cmidrule{12-17}
\textbf{Model} & & \textbf{PEFT} & & B-4 & M & R & C & S & CLIP-S & & B-4 & M & R & C & S & CLIP-S \\
\midrule
CLIP-Captioner~\cite{barraco2022unreasonable} & & - & & 17.3 & 19.5 & 40.8 & 35.7 & 9.4 & 0.650 & & 14.9 & 17.2 & 35.9 & 34.3 & 11.6 & 0.651 \\  
\midrule
\multirow{6}{*}{LLaVA-v1.5-7B~\cite{liu2023improved}} && - & & 15.4 & 16.1 & 40.3 & 41.1 & \underline{\textbf{15.0}} & \underline{\textbf{0.758}} & & 15.0 & \underline{\textbf{18.2}} & 38.5 & 43.5 & \underline{\textbf{19.3}} & \underline{\textbf{0.802}} \\
& & Prompt Learning & & \textbf{20.2} & 15.4 & \textbf{42.5} & \textbf{49.9} & 14.0 & 0.742 & & \textbf{20.5} & 16.2 & 40.4 & 51.5 & 17.0 & 0.758 \\
& & Prefix Tuning & & 19.0 & \underline{\textbf{15.5}} & 41.9 & 47.2 & 14.2 & 0.744 & & 19.6 & 17.0 & \textbf{40.5} & \underline{\textbf{51.7}} & 17.8 & 0.773 \\
& & LoRA & & 19.2 & 14.8 & 42.0 & 43.9 & 13.4 & 0.728 & & 17.1 & 15.0 & 38.0 & 40.2 & 15.2 & 0.730 \\
& & DoRA & & 20.1 & 15.1 & 42.4 & 47.4 & 13.7 & 0.733 & & 18.5 
& 15.5 & 39.0 & 43.5 & 16.0 & 0.738 \\
& & Full Fine-tuning & & 19.5 & 14.7 & 41.8 & 43.7 & 14.7 & 0.725 & & 17.1 & 15.0 & 38.1 & 39.3 & 15.2 & 0.728 \\
\midrule
\multirow{6}{*}{LLaVA-v1.6-7B~\cite{liu2024llavanext}} && - & & 6.9 & 13.0 & 28.6 & 16.9 & 13.7 & 0.725 & & 5.9 & 14.1 & 26.4 & 20.9 & 12.0 & \textbf{0.765} \\
& & Prompt Learning & & \underline{\textbf{20.9}} & \textbf{15.4} & \underline{\textbf{43.0}} & \underline{\textbf{50.4}} & \textbf{14.4} & \textbf{0.740} & & 18.6 & \textbf{15.5} & 39.3 & 44.4 & \textbf{16.5} & 0.739 \\
& & Prefix Tuning & & 14.5 & 13.5 & 36.7 & 35.7 & 11.9 & 0.724 & & 12.9 & 13.8 & 33.8 & 31.8 & 13.8 & 0.737 \\
& & LoRA & & 20.7 & 15.2 & 42.6 & 48.0 & 13.9 & 0.736 & & \underline{\textbf{20.7}} & 15.2 & \underline{\textbf{42.5}} & \textbf{47.8} & 13.8 & 0.737 \\
& & DoRA & & 20.6 & 15.2 & 42.4 & 47.6 & 13.8 & 0.737 & & 18.1 & \textbf{15.5} & 39.0 & 43.7 & 16.0 & 0.746 \\
& & Full Fine-tuning & & 19.8 & 14.8 & 41.9 & 44.1 & 13.3 & 0.721 & & 16.8 & 14.9 & 38.3 & 38.9 & 15.1 & 0.726 \\
\bottomrule
\end{tabular}
}
\vspace{-0.3cm}
\end{table}

\tit{Generalization to Out-of-Domain Settings} Tables~\ref{tab:out-domain-1} and~\ref{tab:out-domain-2} present the results on out-of-domain settings, including nocaps and CC3M datasets (Table~\ref{tab:out-domain-1}) and VizWiz and Textcaps benchmarks (Table~\ref{tab:out-domain-2}). Also in this case, for both LLaVA-v1.5 and LLaVA-v1.6, we compare different fine-tuning strategies with the MLLM tested in zero-shot on the considered datasets and also include the results from the CLIP-Captioner approach. 

As it can be seen, the overall trend is significantly different from the one observed for in-domain evaluation with the standard captioning model trained from scratch on COCO achieving lower results than almost all fine-tuning strategies. The only exception is the CC3M dataset that, however, contains less curated captions than the other datasets, thus leading to less interpretable patterns. Fine-tuning the entire LLM does not lead to the best results in this case, underscoring the need to find viable fine-tuning alternatives to preserve good generalization capabilities in out-of-domain settings. Among the PEFT techniques under consideration, prompt learning is the one achieving the best results on average on all datasets and both LLaVA versions. Similar performances are obtained by LoRA and DoRA fine-tuning strategies, which however fail to preserve high results, especially on the VizWiz dataset.

\begin{figure}[t]
    \centering
    \includegraphics[width=0.98\linewidth]{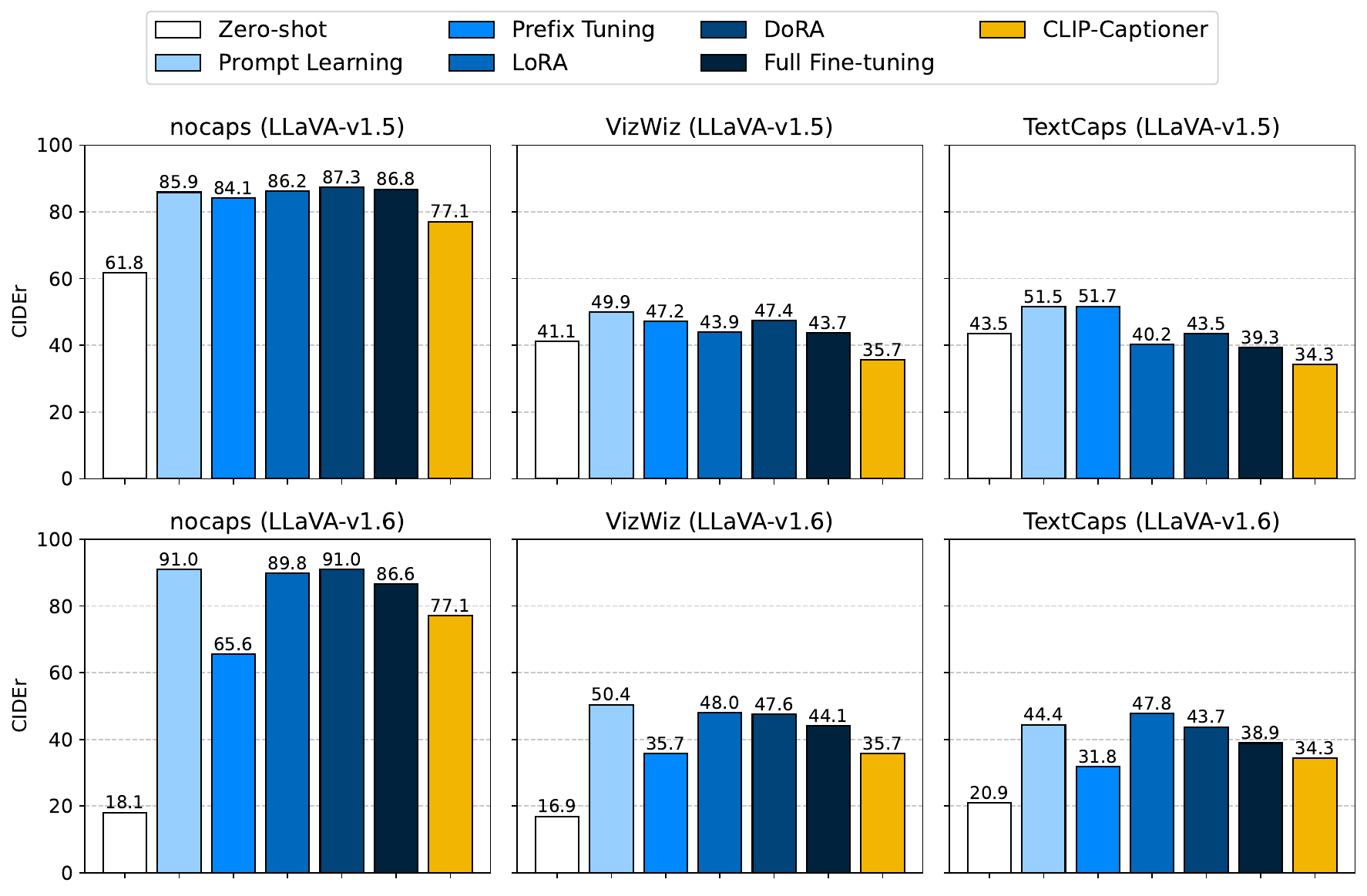}
    \vspace{-0.2cm}
    \caption{Comparison between CIDEr scores achieved by the different versions of LLaVA-v1.5 (first row) and by those of LLaVA-v1.6 (second row) on out-of-domain datasets including nocaps, VizWiz, and TextCaps.}
    \label{fig:plots}
    \vspace{-0.3cm}
\end{figure}

Also in these settings, captions generated by zero-shot MLLMs are confirmed to be far from ground-truth image descriptions contained in each considered dataset, as demonstrated by the low scores in terms of standard captioning metrics achieved by these models. These results highlight the need of proper fine-tuning strategies to adapt MLLMs for the task of image captioning and the necessity of novel evaluation protocols that take into account the different descriptive styles of the textual descriptions generated by these models.

A different visualization of the results is shown in Fig.~\ref{fig:plots} where we compare the CIDEr scores of the considered models on nocaps, VizWiz, and TextCaps. Notably, the scores achieved by the zero-shot MLLMs are always below all other fine-tuned versions. This is particularly evident with LLaVA-v1.6 which tends to generate longer captions and is more prone to hallucinations. Overall, prompt learning better generalizes across different out-of-domain datasets, always achieving the best or second-best results in almost all settings and considering both LLaVA-v1.5 and LLaVA-v1.6.

\begin{figure}[t]
    \begin{minipage}{0.165\linewidth}
        \includegraphics[width=0.97\linewidth]{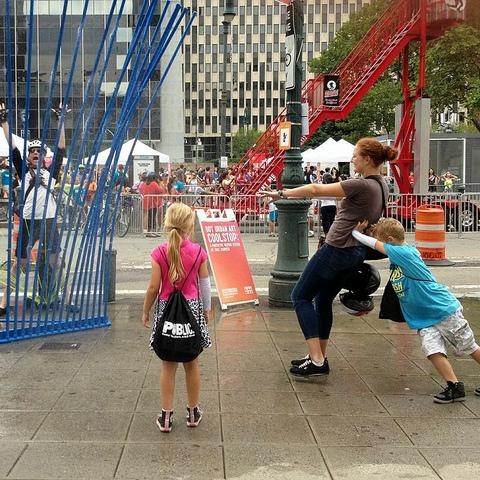}
        \end{minipage}
        \begin{minipage}{0.32\linewidth}
        \scriptsize{
        \textbf{GT:} A girl in a pink shirt standing near a blue metal sculpture.\\
        \textbf{Zero-shot:} A woman and children are standing in front of a blue fence, with the woman holding a child. They are near a pole and a sign.\\
        \textbf{Prompt learning:} A woman and children are posing for a picture.
        }
    \end{minipage}
    \hspace{0.02cm}
    \begin{minipage}{0.165\linewidth}
     \includegraphics[width=0.97\linewidth]{images/qualitatives/basket.jpg}
        \end{minipage}
        \begin{minipage}{0.32\linewidth}
        \scriptsize{
        \textbf{GT:} The TV screen has a man who is reporting live.\\
        \textbf{Zero-shot:} A man is standing in front of a TV, which is displaying a basketball game. The TV screen shows the score and the teams playing.\\
        \textbf{Prompt learning:} A man on a television screen talking about the Rockets and the Grizzlies.
        }
    \end{minipage}
    
    \vspace{0.15cm}

    \begin{minipage}{0.165\linewidth}
    \includegraphics[width=0.97\linewidth]{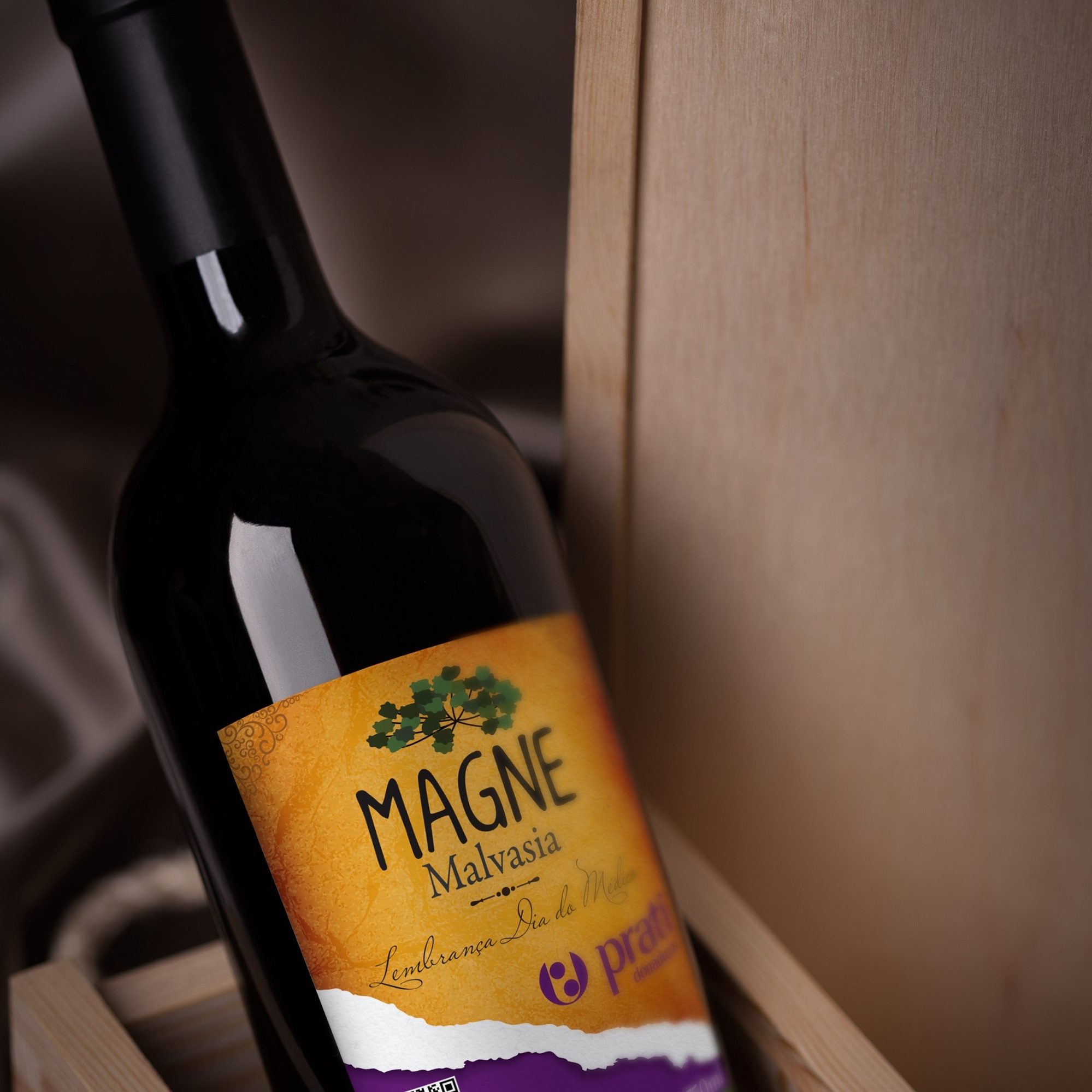}
        \end{minipage}
        \begin{minipage}{0.32\linewidth}
        \scriptsize{
        \textbf{GT:} A bottle of wine is labeled with the name MAGNE.\\
        \textbf{Zero-shot:} A bottle of wine is placed in a wooden box, with the label showing that it is a 2013 vintage.\\
        \textbf{Prompt learning:} A bottle of wine with a label that says Magne.
        }
    \end{minipage}
    \hspace{0.02cm}
    \begin{minipage}{0.165\linewidth}
       \includegraphics[width=0.97\linewidth]{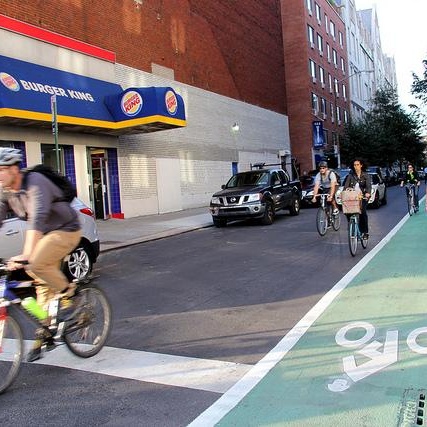}
        \end{minipage}
        \begin{minipage}{0.32\linewidth}
        \scriptsize{
        \textbf{GT:} Bike riders passing Burger King in city street.\\
        \textbf{Zero-shot:} A group of people riding bicycles down a street, with a man on a bike in front of a Subway sandwich shop.\\
        \textbf{Prompt learning:} A group of people riding bikes on a city street.
        }
    \end{minipage}

    \vspace{0.15cm}

    \begin{minipage}{0.165\linewidth}
        \includegraphics[width=0.97\linewidth]{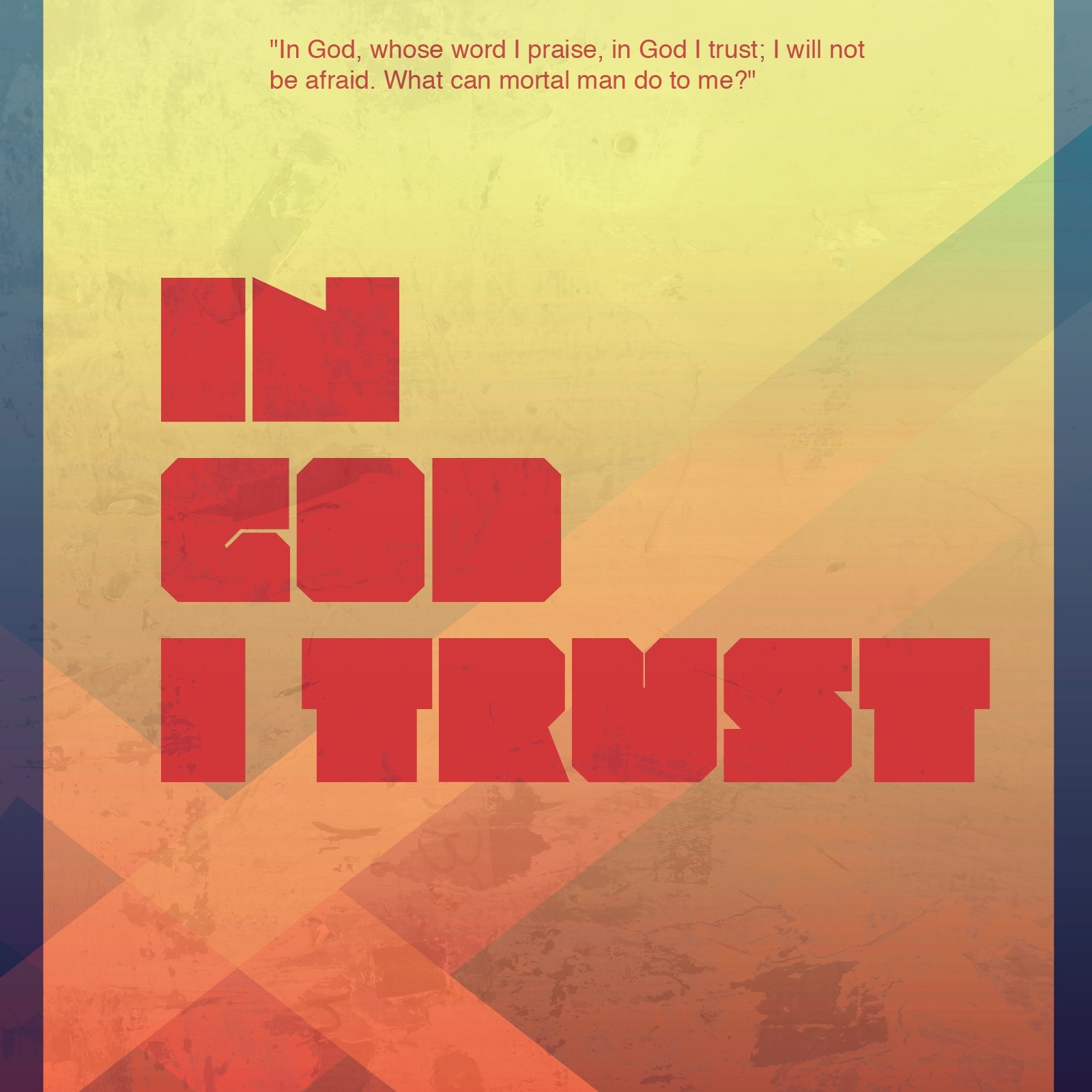}
        \end{minipage}
        \begin{minipage}{0.32\linewidth}
        \scriptsize{
        \textbf{GT:} A poster for the book named in god I trust.\\
        \textbf{Zero-shot:} Colorful poster featuring Psalm 46:10, ``Be still, and know that I am God''.\\
        \textbf{Prompt learning:} A poster that says ``In God I Trust''.
        }
    \end{minipage}
    \hspace{0.02cm}
    \begin{minipage}{0.165\linewidth}
        \includegraphics[width=0.97\linewidth]{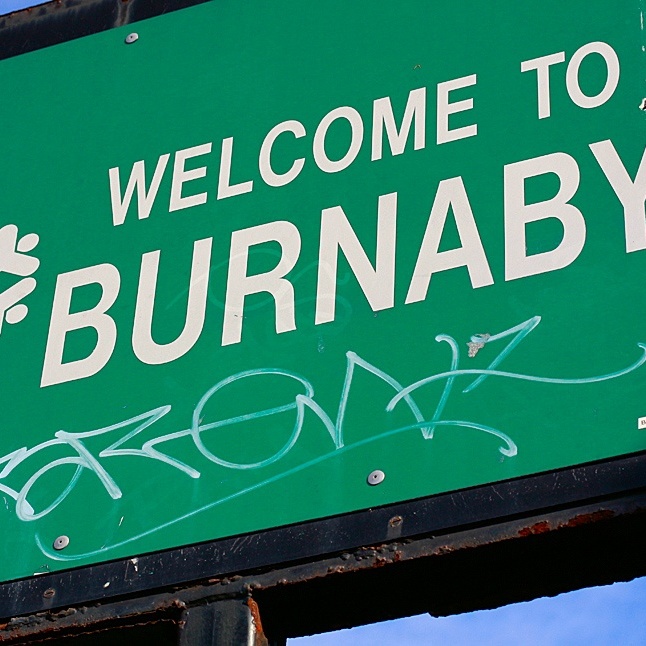}
        \end{minipage}
        \begin{minipage}{0.32\linewidth}
        \scriptsize{
        \textbf{GT:} A sign that is green and says ``Welcome to Burnaby''.\\
        \textbf{Zero-shot:} Welcome to Burnaby: A city where snowflakes are a symbol of the season, and graffiti is a common sight.\\
        \textbf{Prompt learning:} A sign that says ``Welcome to Burnaby`` with graffiti on it.
        }
    \end{minipage}
\vspace{-.2cm}
\caption{Sample image descriptions generated by the zero-shot MLLM in comparison with those generated by the MLLM fine-tuned with prompt learning. For reference, we also report a ground-truth caption (GT) associated to each image.}
\label{fig:qualitatives}
\vspace{-.3cm}
\end{figure}

\begin{figure}[t]
    \begin{minipage}{0.165\linewidth}
        \includegraphics[width=0.97\linewidth]{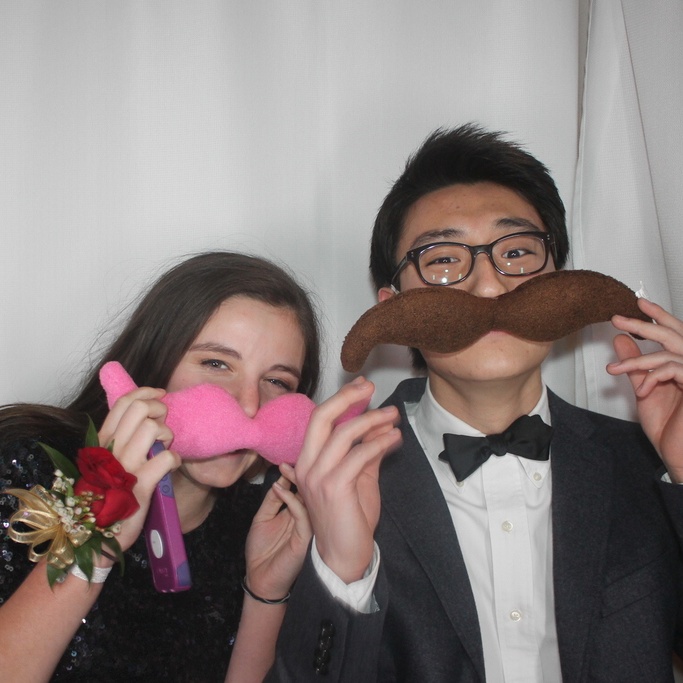}
        \end{minipage}
        \begin{minipage}{0.32\linewidth}
        \scriptsize{
        \textbf{CLIP-Captioner:} A man with mustaches.\\
        \textbf{Full fine-tuning:} A man and a woman holding up pink and brown objects.\\
        \textbf{Prompt learning:} A woman and a man with fake mustaches on their mouths.
        }
    \end{minipage}
    \hspace{0.02cm}
    \begin{minipage}{0.165\linewidth}
        \includegraphics[width=0.97\linewidth]{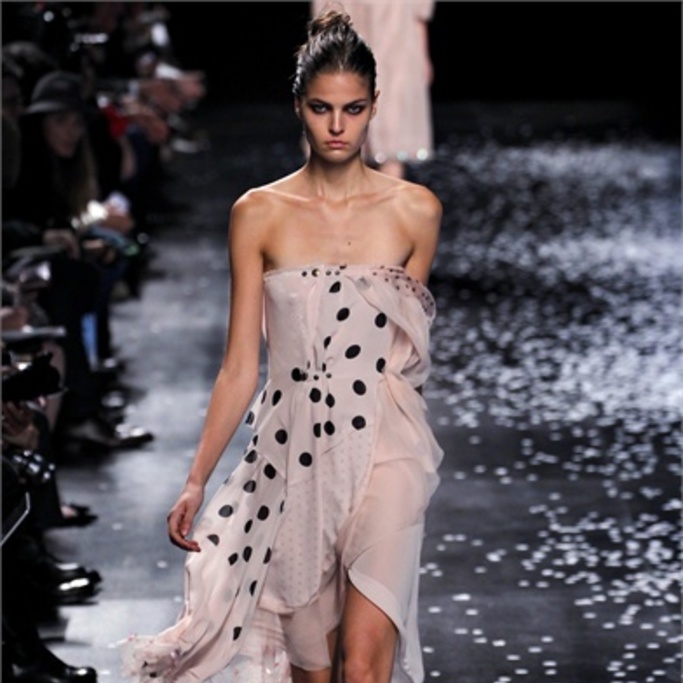}
        \end{minipage}
        \begin{minipage}{0.32\linewidth}
        \scriptsize{
        \textbf{CLIP-Captioner:} A woman in uniform standing in front of a computer screen.\\
        \textbf{Full fine-tuning:} A woman is walking down a runway with a dress on.\\
        \textbf{Prompt learning:} A woman walking down a runway in a dress with polka dots.
        }
    \end{minipage}

    \vspace{0.1cm}

    \begin{minipage}{0.165\linewidth}
        \includegraphics[width=0.97\linewidth]{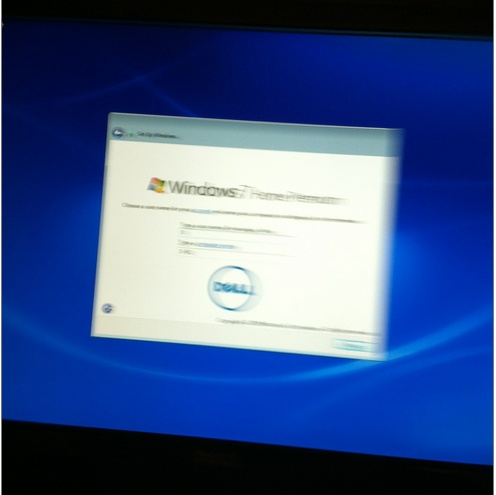}
        \end{minipage}
        \begin{minipage}{0.32\linewidth}
        \scriptsize{
        \textbf{CLIP-Captioner:} A screenshot of a computer screen.\\
        \textbf{Full fine-tuning:} A computer screen with a Windows logo on it.\\
        \textbf{Prompt learning:} A computer screen with a blue background and a Windows error message.
        }
    \end{minipage}
    \hspace{0.02cm}
    \begin{minipage}{0.165\linewidth}
        \includegraphics[width=0.97\linewidth]{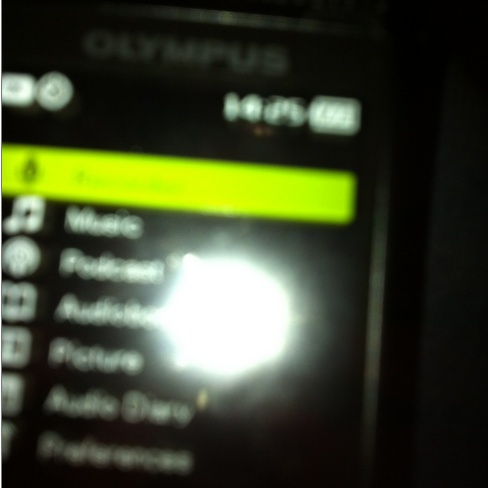}
        \end{minipage}
        \begin{minipage}{0.32\linewidth}
        \scriptsize{
        \textbf{CLIP-Captioner:} A screenshot of my computer.\\
        \textbf{Full fine-tuning:} A digital music player with a yellow menu.\\
        \textbf{Prompt learning:} A phone screen with a music app and a podcast app.
        }
    \end{minipage}
    \vspace{0.1cm}

    \begin{minipage}{0.165\linewidth}
        \includegraphics[width=0.97\linewidth]{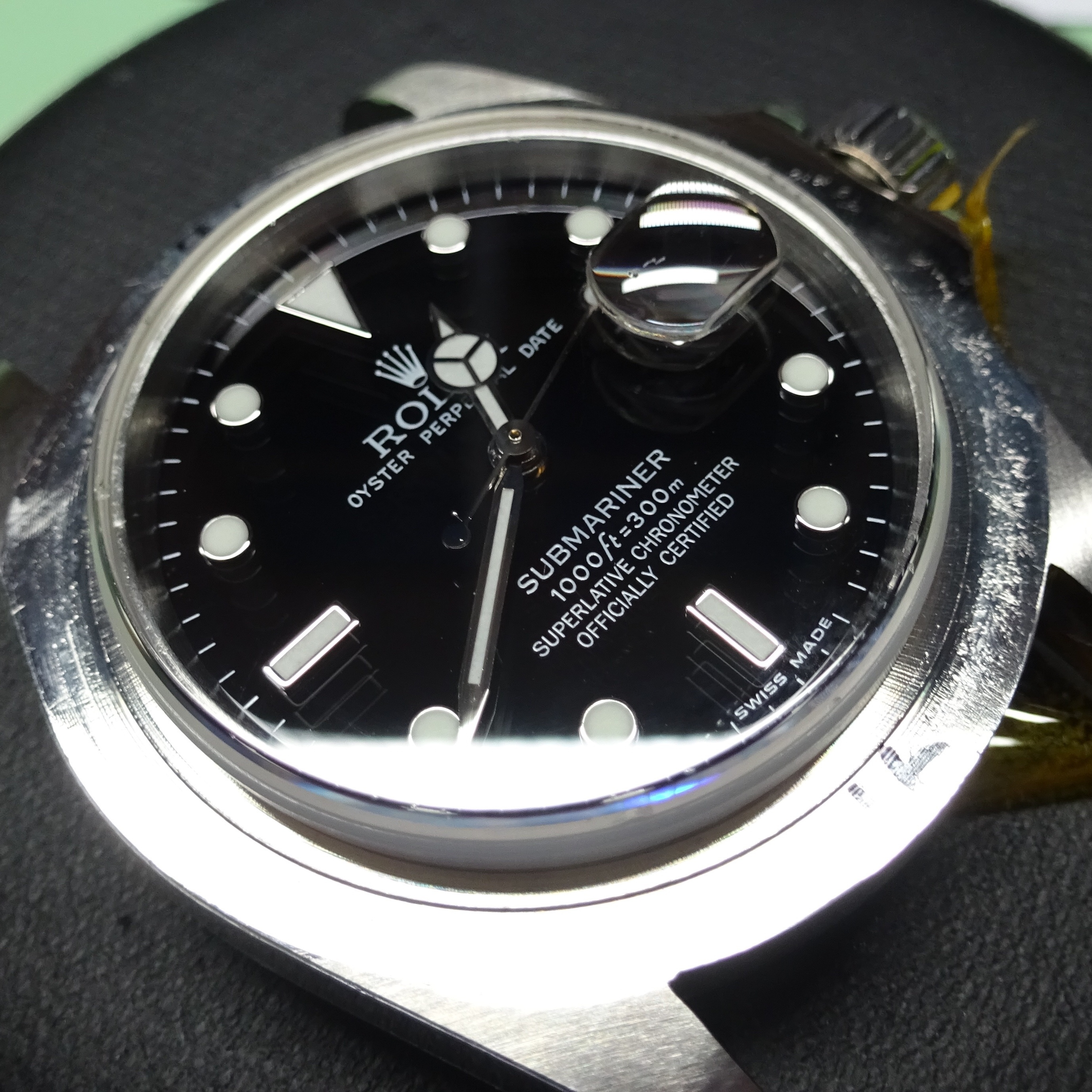}
        \end{minipage}
        \begin{minipage}{0.32\linewidth}
        \scriptsize{
        \textbf{CLIP-Captioner:} A watch on a table.\\
        \textbf{Full fine-tuning:} A watch sitting on a table with a broken glass.\\
        \textbf{Prompt learning:} A Rolex watch is laying on a table.
        }
    \end{minipage}
    \hspace{0.02cm}
    \begin{minipage}{0.165\linewidth}
        \includegraphics[width=0.97\linewidth]{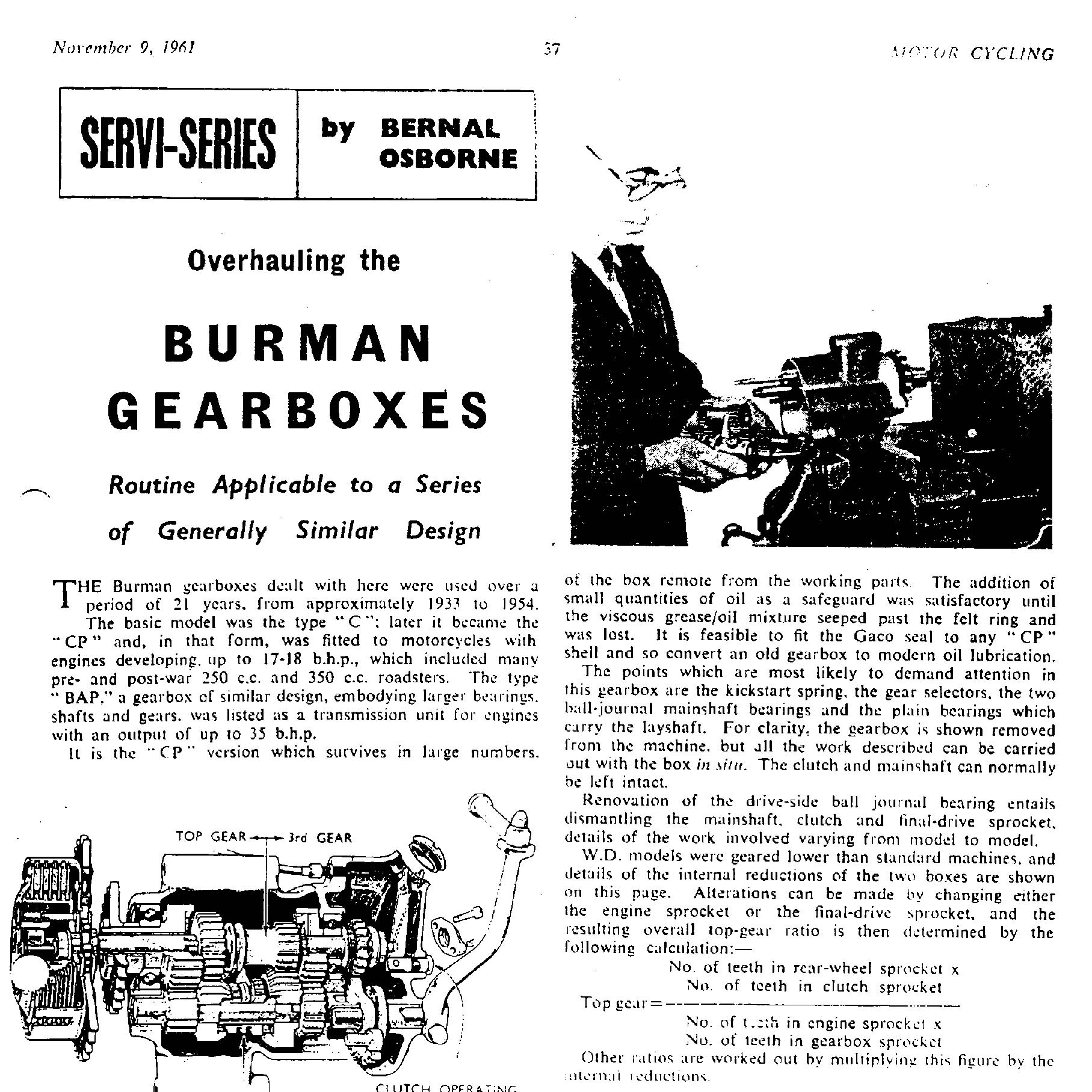}
        \end{minipage}
        \begin{minipage}{0.32\linewidth}
        \scriptsize{
        \textbf{CLIP-Captioner:} A black and white page.\\
        \textbf{Full fine-tuning:} A page of a book with a picture of a machine.\\
        \textbf{Prompt learning:} A page from a magazine that is about Burman gearboxes.
        }
    \end{minipage}
\vspace{-.2cm}
\caption{Qualitative results on sample images from nocaps (first row), ViZWiz (second row), and TextCaps (third row). We compare captions predicted by the CLIP-Captioner model~\cite{barraco2022unreasonable}, the MLLM after full fine-tuning, and the MLLM after prompt learning.}
\label{fig:qualitatives2}
\vspace{-.3cm}
\end{figure}

\tit{Qualitative Results} Finally, we report some qualitative results in Fig.~\ref{fig:qualitatives} and~\ref{fig:qualitatives2}. Specifically, in Fig.~\ref{fig:qualitatives} we compare captions generated by the zero-shot MLLM (both LLaVA-v1.5 and LLaVA-v1.6) with those generated by the MLLM fine-tuned with prompt learning, which demonstrates to be one of the best fine-tuning solutions for the image captioning task. For completeness, we also include a sample ground-truth caption. Notably, while captions generated by the zero-shot MLLM are generally longer and more detailed, they often contain hallucinations or fail to well describe the visual content of the input image. For example, in the second row-left sample, the zero-shot MLLM correctly identifies the bottle of wine but also reports the vintage year which however is not shown in the image. Similarly, in the third row-right sample, the zero-shot MLLM correctly reads the text written in the image but provides details on the city whose name appears in the written text. These additional details, however, do not help to better describe the visual content appearing in the scene. In both cases, instead, the fine-tuned version of the MLLM can generate a concise caption, while still describing the key concepts depicted in the images. 

In Fig.~\ref{fig:qualitatives2}, we report additional qualitative results, in this case comparing the MLLM fine-tuned with prompt learning with the predictions generated by the CLIP-Captioner approach and those generated by the MLLM after a full fine-tuning stage. As it can be seen, fine-tuning the MLLM with a PEFT-based solution leads to captions that are generally more detailed, while still preserving the concise and timely style of standard captioning benchmarks. On the contrary, training from scratch a captioning model on COCO or directly optimizing all MLLM parameters causes a loss of generality, especially when the model should describe objects or concepts that are not present in the training dataset. These results confirm from a qualitative point of view the effectiveness of using appropriate fine-tuning strategies to adapt an existing MLLM to the image captioning task and show that utilizing a full fine-tuning of the model is not the preferable choice in this setting.

\subsection{Computational Analysis}
Finally, in Table~\ref{tab:efficiency} we present a computational and energy consumption analysis for the fine-tuning strategies under consideration on LLaVA-v1.5. For each PEFT strategy, we report the number of trainable parameters along with the energy consumed during training, measured in Kilowatt-hours (kWh). Energy consumption is detailed both for the entire training process on the COCO training split (\ie~four epochs in our experiments) and for the epochs up to the best checkpoint, selected based on validation loss. As it can be seen, prompt learning and prefix tuning are the least computationally demanding strategies. Furthermore, while training with LoRA or DoRA consumes a similar amount of energy as full fine-tuning when considering all epochs, DoRA generally converges in fewer iterations, leading to lower overall energy consumption.

\begin{table}[t]
\caption{Computational analysis in terms of trainable parameters and energy consumed during training. Energy consumption is reported for the entire training process as well as for the epochs up to the best checkpoint (with the latter shown in parentheses). All experiments have been conducted on four 64GB NVIDIA A100 GPUs.}
\vspace{-0.1cm}
\label{tab:efficiency}
\footnotesize
\centering
\setlength{\tabcolsep}{.3em}
\resizebox{0.75\linewidth}{!}{
\begin{tabular}{lccc cc cc}
\toprule
& & & & \textbf{Trainable} & & \textbf{Energy}  \\
\textbf{Model} & & \textbf{PEFT} & & \textbf{Params} & & \textbf{Consumption (kWh)}  \\
\midrule
\multirow{5}{*}{LLaVA-v1.5-7B~\cite{liu2023improved}} 
& & Prompt Learning & & 19.7k & & 12.4 (3.1)\\
& & Prefix Tuning & & 13.1k & & 12.1 (3.0) \\
& & LoRA & & 319.8M & & 59.5 (28.5) \\
& & DoRA & & 321.2M & & 58.2 (14.5)\\
& & Full Fine-tuning & & 7B & & 64.3 (32.1) \\
\bottomrule
\end{tabular}
}
\vspace{-0.3cm}
\end{table}

\section{Conclusion}
\label{sec:conclusion}
This paper has explored the intersection of image captioning and the rapidly evolving landscape of Multimodal LLMs, assessing their potential as effective replacements for specialized image captioning networks. Through comprehensive experiments and analyses across multiple image description benchmarks, we have demonstrated the limitations of common MLLMs when applied to this task without specific training. In fact, captions generated by standard MLLMs are often prone to hallucinations and struggle to adhere to the concise, grammatically correct, and object-focused description style characteristic of standard image captioning datasets. While standard fine-tuning schemes can improve performance to some extent, they often come at the cost of reduced generalization. To bridge this gap, we have analyzed the effectiveness of various PEFT techniques for adapting MLLMs to the image captioning task, showing that these solutions generally yield better results in terms of both coherence with ground-truth captions and generalization to out-of-domain settings. Our findings suggest the necessity for further research to design effective strategies for adapting existing MLLMs in this domain, mainly focusing on improving their ability to generate accurate, concise, and hallucination-free image captions while maintaining generalization across different domains.

\section*{Acknowledgments}
We acknowledge the CINECA award under the ISCRA initiative, for the availability of high-performance computing resources. This work has been conducted under a research grant co-funded by Altilia s.r.l. and supported by the PNRR-M4C2 (PE00000013) project ``FAIR - Future Artificial Intelligence Research'' and by the PRIN 2022-PNRR  M4C2-I1.1 project ``MUCES - a MUltimedia platform for Content Enrichment and Search in audiovisual archives'' (CUP E53D23016290001), both funded by EU - Next-Generation EU.


\bibliographystyle{splncs04}
\bibliography{bibliography}
\end{document}